# Fine-Tuning Approach for Arabic Offensive Language Detection System

## BERT-Based Model

Fatemah Husain
Department of Information Science, Kuwait University
Kuwait City, State of Kuwait
f.husain@ku.edu.kw

Ozlem Uzuner
Department of Information Sciences and Technology
George Mason University
Fairfax, USA
ouzuner@gmu.edu

*Abstract*—The problem of online offensive language limits the health and security of online users. It is essential to apply the latest state-of-the-art techniques in developing a system to detect online offensive language and to ensure social justice to the online communities. Our study investigates the effects of fine-tuning across several Arabic offensive language datasets. We develop multiple classifiers that use four datasets individually and in combination in order to gain knowledge about online Arabic offensive content and classify users' comments accordingly. Our results demonstrate the limited effects of transfer learning on the classifiers' performance, particularly for highly dialectal comments.

## I. INTRODUCTION

Developing a system to detect online offensive language is very important to the health and the security of online users. Studies have shown that cyberhate, online harassment, and other misuses of technology are on the rise. Particularly during the global Coronavirus pandemic in 2020, 35% of online users reported online harassment related to their identity-based characteristics, reflecting a 3% increase over 2019[1].

Applying advanced techniques from Natural Language Processing (NLP) to support the development of an online hate-free community is a critical task for social justice.

This study aims at investigating the effects of fine-tuning a Bidirectional Encoder Representations from Transformers (BERT) model on multiple Arabic offensive language datasets individually and testing it using other datasets individually. Our experiment starts with a comparison among multiple BERT models to guide the selection of the main model for our study. The study also investigates the effects of concatenating all datasets for fine-tuning BERT.

## II. BACKGROWND

### A. Arabic Language

This study covers Arabic text from online user-generated content. While there are multiple forms of the Arabic language, the majority of the content from user-generated platforms is written in dialectal Arabic. The dialectal form of Arabic is the actual spoken Arabic, and it has several categories depending on social and geographical factors. Habash [1] divides the Arabic dialects into seven categories; Egyptian, Levantine, Gulf, North African, Iraqi, Yemenite, and Maltese. The diversity among Arabic dialects poses difficulties to NLP systems aimed at processing online Arabic content.

### B. Offensive Language

Offensive language refers to include abusive statements that aim to harm, such as threats, discriminatory words, swear words, blunt insults, hate speech, aggressive content, cyberbullying, and toxic comments [2; 3]. Offensive exchanges create an environment of disturbance, disrespect, anger, affecting the harmony of conversations, and reduces users' trust in the online platforms [4].

### C. Related Work

BERT is an innovative language model that presents state-of-the-art results in multiple NLP tasks, such as question answering and language inference. BERT applies pre-trained language representations to down-stream tasks through a fine-tuning process. The main feature that distinguishes BERT from the other language representations is the use of a bidirectional language model (rather than unidirectional) during pre-training. BERT's bidirectional language model is a Masked Language Model (MLM), which randomly masks some of the tokens from the input with the objective of predicting the original token of the masked word based only on its context [5].

---

[1] https://www.adl.org/

Multilingual BERT (M-BERT)[2], proposed by Google Research, has two versions; BERT-base-multilingual-uncased which covers 102 languages and BERT-base-multilingual-cased, which covers 104 languages. Wikipedia dumps of each language (excluding user and talk pages) were used to train the models with a shared word piece vocabulary. Previous studies report that M-BERT outperforms other tools on multilingual text. However, M-BERT shows some limitations in tokenizing Arabic sentences [6]. This finding is in line with other experiments conducted by Hasan et al. [7], Saeed et al. [8], and Keleg et al. [9], who report poor performance in Arabic offensive language and hate speech detection compared to other word embeddings.

In addition, Abu Farha and Magdy [10] try M-BERT with Adam optimizer, and fine-tune the model with 4 epochs, learning rate of 1e−5, and setting the maximum sequence length to the maximum length seen in the training set; the results were not as good as the results obtained from the Bidirectional Long Short-Term Memory (BiLSTM) model, Convolutional Neural Network - Bidirectional Long Short-Term Memory (CNN-BiLSTM) model, and multitask learning models. In [11], multiple M-BERT-based classifiers were used with different fine-tuning settings for offensive language and hate speech detection tasks, and in both tasks the reported macro-averaged F1 score was not better than what has been reported by other studies using simple traditional machine learning methods [12].

While M-BERT supports various languages, Arabic specific BERT models have been used as well for Arabic offensive language detection, such as AraBERT and BERT-Arabic. AraBERT[3] is an Arabic version of BERT that shows state-of-the-art performance in multiple downstream tasks [13]. It uses the BERT-base configuration, and has similar pre-training settings as the ones used by the original BERT model, consisting of a MLM and Next Sentence Prediction (NSP) task. Multiple Modern Standard Arabic (MSA) corpora are used to train the model, which include: manually scraped Arabic news websites for articles, 1.5 billion words extracted from news articles from ten major news sources, and OSIAN, which is an Open Source International Arabic News Corpus. AraBERT outperforms both M-BERT and other state-of-the-art models on sentiment analysis, question answering, and Named Entity Recognition (NER) [14, 6]. This finding demonstrates that a pre-trained language model trained on a single language performs better than a multilingual model when applied in a monolingual text.

Djandji et al.[15] apply AraBERT to the Open-Source Arabic Corpora and Corpora Processing Tools (OSACT) dataset for multitask vs. multilabel classification. Multitask Learning solves the data imbalance problem in OSACT dataset by leveraging information from multiple tasks simultaneously. The same study also applies AraBERT for multilabel classification, in which all labels of the two labeling hierarchies in OSACT dataset—offensive and hate—are merged under a broad task of violence detection. Results report 90.15% as the highest macro-averaged F1 score for offensive language detection using multitask learning. Findings from this study demonstrate the superiority of multitask learning over multilabel classification using AraBERT for offensive language detection. The error analysis reveals that confusion occurs in tweets that consist of offensive words in a non-offensive context. It also shows that most of the errors are related to mockery, sarcasm, or mentioning other offensive and hateful statements within tweets.

Arabic-BERT[4] is another Arabic monolingual BERT model [16]. The pre-trained corpus consists of multiple Arabic resources such as Arabic OSCAR and Arabic Wikipedia, which includes MSA and dialectal Arabic. Results from evaluating the model's performance for sentiment analysis shows higher F1 score for Arabic-BERT than M-BERT, and the hULMonA[5] model (previous state-of-the-art model for Arabic sentiment analysis) when used with Levantine dialect and Egyptian dialect datasets.

The work in this paper buids upon findings from previous studies. It evaluates four BERT models for Arabic offensive language detection, and further investigates the AraBERT model under various fine-tuning settings using four datasets for Arabic offensive language detection from different platforms and domains.

III. METHODOLOGY

A. Datasets

We use four publicly available Arabic offensive language datasets. These datasets include: Aljazeera.net Deleted Comments [17], YouTube dataset [18], Levantine Twitter Dataset for Hate Speech and Abusive Language (L-HSAB) [19], and the OSACT offensive and not offensive classification samples [20]. Table 1 provides a summary for the characteristics of each dataset.

We use only binary classes; offensive or not offensive. Thus, we convert different types of offensive languages to offensive class. For example, the L-HSAB dataset differentiates between hate and abusive language classes; which were both converted to offensive language class.

---

[2] https://github.com/google-research/bert/blob/master/multilingual.md
[3] https://github.com/aub-mind/arabert
[4] https://github.com/alisafaya
[5] https://github.com/aub-mind/hULMonA

For some datasets that are provided in a train/evaluation/test splitted formats, we merge all parts together into one dataset, and then, we randomly apply 80%-20% split for train-test datasets. This supports consistency among the datasets used in this study as most of them are provided in one part. All datasets were used without any preprocessing.

TABLE I. DATASET DESCRIBTION

| Dataset | Source | Labels/Size |
|---|---|---|
| Aljazeera Deleted Comments (Mubarak et al., 2017) | Aljazeera News (Aljazeera.net) | 31,692 comments (offensive = 25,506, clean = 5,653, obscene = 533) |
| YouTube Comments (Alakrot, Murray, & Nikolov, 2018) | YouTube | 15,050 comments (not offensive = 9,237, offensive = 5,813) |
| Levantine Twitter Dataset for Hate Speech and Abusive Language (L-HSAB) (Mulki et al., 2019) | Twitter | 5,846 tweets (hate = 468, abusive = 1,728, normal = 3,650) |
| Open-Source Arabic Corpora and Corpora Processing Tools (OSACT) (Mubarak et al., 2020) | Twitter | 10,000 tweets (offensive = 1,900, not offensive = 8,100) (hate = 500, not hate = 9,500) |

### B. The BERT Model

Our experiments depend mainly on the AraBERT model from the Huggingface[6] library. To select the best available BERT model for our task, we evaluate the available models on the OSACT and the L-HSAB datasets by training the models on 80% of the training datasets and applying the trained models to 20% of the remaining training datasets. The evaluated models include XLM-Roberta[7] (also called XLM-R), M-BERT, Arabic-BERT, and AraBERT. Table II shows the macro-averaged F1 scores for each of the models. As can be noticed from Table II, AraBERT outperforms all other models on both datasets. Thus, we use AraBERT for our main experiments. Moreover, the table also demonstrates that Arabic monolingual models perform better than multilingual models.

In all experiments, including Table II models evaluation, we apply the same experiment settings: maximum length = 128, patch size = 16, epoch = 5, epsilon = 1e-8, and learning rate = 2e-5. We use the pooled output from the encoder with a simple Feed Forward Neural Network (FFNN) layer to build the classifier. Experiments were developed in Python using PyTorch-Transformers library, and evaluation metrics were developed using Scikit-Learn Python library. Google Colab Pro used to conduct all experiments.

TABLE II. MACRO-AVERAGED F1 SCORES FROM THE PRELIMINARY STUDIES OF DIFFERENT BERT MODELS

| Dataset | Arabic Monolingual Model | | Multilingual Model | |
|---|---|---|---|---|
| | AraBERT | Arabic-BERT | M-BERT | XLM-RoBERTa |
| OSACT-Offensive | **0.90** | 0.88 | 0.87 | 0.86 |
| L-HSAB | **0.88** | 0.85 | 0.83 | 0.79 |

Given the choice of AraBERT, we evalaute it on four datasets. For this purpose, we fine-tuned the pre-trained AraBERT model to the training portion of each dataset and apply it to the test portion of the same dataset.

## IV. RESULTS AND DISCUSSION

### A. Results

We use macro-averaged precision, recall, F1, and accuracy to evaluate the classifiers' performance. The following table shows results for the four individual models, each fine-tuned using the training portion of a dataset and tested on the test set.

TABLE III. PERFORMANCE RESULTS

| Fine-Tuned and Tested Dataset | Precision | Recall | F1 | Accuracy |
|---|---|---|---|---|
| OSACT | **0.91** | **0.91** | **0.91** | **0.94** |
| L-HSAB | 0.87 | 0.87 | 0.87 | 0.88 |
| YouTube | 0.87 | 0.87 | 0.87 | 0.88 |
| Aljazeera | 0.73 | 0.75 | 0.74 | 0.85 |

As can be noticed from the table above, the highest recorded macro-averaged recall, macro-averaged F1, and accuracy scores are shown for the OSACT dataset. Aljazeera dataset shows lowest overall performance scores.

Table IV shows the results after concatenating all training datasets into one corpus, and using it to fine-tune the classifier. Overall, the table shows no gains over Table III. The OSACT dataset is still recording the same highest performance. However, the L-HSAB dataset shows a reduction in performance by 3% in macro-averaged F1 score. This decrease could be a result of the high dialectal text of the L-HSAB, as other datasets might not contain much Levantine vocabulary words.

TABLE IV. PERFORMANCE RESULTS FROM THE CONCATENATED FINE-TUNED MODEL

| Testing Dataset | Precision | Recall | F1 | Accuracy |
|---|---|---|---|---|
| OSACT | **0.90** | **0.91** | **0.90** | **0.94** |
| L-HSAB | 0.86 | 0.84 | 0.84 | 0.85 |
| YouTube | 0.87 | 0.87 | 0.87 | 0.88 |
| Aljazeera | 0.74 | 0.75 | 0.74 | 0.85 |

---

[6] https://huggingface.co/
[7] https://github.com/pytorch/fairseq/tree/master/examples/xlmr

## B. Error Analysis

We calculate percentages of misclassified comments per class for each experiment conducted using the concatenated fine-tuned model. Table V presents a summary of the error analysis. The percentages are calculated per class based on the total instances of each class for each of the four testing datasets. Most common five tokens are presented.

As can be seen from table V, offensive and not offensive misclassified percentages vary among the datasets. Top tokens among the misclassified samples include names of countries (e.g., Saudi Arabia, Qatar, Iraq) and names of famous people (e.g., Kadim, Ahlam, Gibran), which in some cases refer to the first name and the last name of the same person in two separate tokens. For example, 'Kadim' and 'Sahir' are the first and the last name of the same singer. Preprocessing procedures can ensure proper understanding of multiple token terms and compound nouns.

TABLE V. ERROR ANALYSIS SUMMARY FOR THE CONCATENATED FINE-TUNED MODEL

| Test Dataset | Misclassified % | | Misclassified Top Common Tokens | |
|---|---|---|---|---|
| | Offensive | Not Offensive | Offensive | Not Offensive |
| OSACT | 16% | 3.5% | الله/God<br>علوقيه/stereotype<br>صغير/small<br>بنت/girl<br>حبيبي/lovely | الله/God<br>عيون/eyes<br>ناس/people<br>موضوع/topic |
| L-HSAB | 8.7% | 19.4% | جبران/Gibran<br>باسيل/Basil<br>حلوة/beautiful<br>وزير/minister<br>نعيم/blessing | جبران/Gibran<br>باسيل/Basil<br>سوريا/Syria<br>قطر/Qatar<br>دولة/state |
| YouTube | 15% | 11% | كاظم/Kadim<br>الله/God<br>احلام/Ahlam<br>ساهر/Sahir<br>عراقي/Iraqi | كاظم/Kadim<br>الله/God<br>ناس/people<br>مصر/Egypt<br>ساهر/Sahir |
| Aljazeera | 8.2% | 44.7% | الله/God<br>دولة/state<br>عراق/Iraq<br>شعب/nation<br>جزيرة/Jazeera | الله/God<br>جزيرة/Jazeera<br>سعودية/Saudi Arabia<br>دولة/state<br>مسلمين/Muslims |

Since our system depends on the vocabulary list from the AraBERT model, we further investigate the AraBERT training corpus. AraBERT is trained mostly on Arabic News outlets, the following are the sources of its training raw text:

1. Arabic Wikipedia database dump[8]
2. 1.5B words Arabic Corpus [9] (sources include newspapers, books, and research papers)
3. Open Source International Arabic News Corpus (OSIAN)[10]
4. Assafir Lebanese news articles[11]
5. Manually crawled news websites (Al-Akhbar, Annahar, AL-Ahram, and AL-Wafd).

Our datasets consist mostly of user-generated content, which differ from the Arabic text that is used in writing news articles and books [21]. The type of Arabic that is used in our datasets is the dialectal Arabic, while the one used in training AraBERT is the MSA. Thus, simple fine-tuning process might not be enough to adjust the weights of AraBERT vocabulary toward our task of offensive language detection, especially if most of the tokens in our datasets are treated as out-of-vocabulary tokens by AraBERT tokenizer.

## C. System Implications

Increasing the dataset size for fine-tuning AraBERT model might not always improve the system's performance. Thus, finding some other methods to improve the performance are required. For example, we may create a more advanced classifier architecture on top of the BERT model that gives better results than those obtained from a simple FFNN classifier. Another method could focus on AraBERT model and trying to adjust its vocabulary to support offensive language classification task. A costlier approach could be to consider training a new BERT model that is customized for the online Arabic offensive language detection task.

## V. CONCLUSION

In this paper, we try to present our work applying AraBERT to several Arabic offensive language detection datasets. Our results report outperformance of Arabic monolingual BERT models over BERT multilingual models. While the results from aggregating knowledge from multiple datasets on the same time show no effects on the performance when tested on individual datasets, it lowers the performance of the highly dialectal dataset; L-HSAB; by 3% in macro-averaged F1 score. The overall findings from our experiments demonstrate the importance of developing novel methods to fine-tune the BERT model for the Arabic offensive language detection tasks.

---

[8] https://archive.org/details/arwiki-20190201
[9] https://arxiv.org/pdf/1611.04033.pdf
[10] https://www.aclweb.org/anthology/W19-4619/
[11] http://m.assafir.com/Channel/50/English/TopMenu